\documentclass{article}
\usepackage{microtype}
\usepackage{graphicx}
\usepackage{booktabs}
\usepackage{hyperref}

\usepackage[preprint]{icml2027}
\renewcommand{\today}{November 27, 2026}
\hypersetup{pdfsubject={Preprint}}
\usepackage{amsmath,amssymb,mathtools,amsthm}
\usepackage{algorithm,algorithmic}
\usepackage[capitalize,noabbrev]{cleveref}

\newtheorem{proposition}{Proposition}

\newcommand{\Z}{\mathbf Z}
\newcommand{\U}{\mathbf U}
\newcommand{\A}{\mathbf A}
\newcommand{\C}{\mathbf C}
\newcommand{\D}{\mathbf D}
\newcommand{\M}{\mathbf M}
\newcommand{\I}{\mathbf I}
\newcommand{\Pim}{\boldsymbol\Pi}
\newcommand{\pik}{\boldsymbol\pi_k}
\newcommand{\partn}{\mathcal B}
\newcommand{\R}{\mathbb R}
\DeclareMathOperator{\Diag}{Diag}
\DeclareMathOperator{\BlockDiag}{BlockDiag}
\DeclareMathOperator{\tr}{tr}
\DeclareMathOperator{\range}{range}

\DeclareMathOperator{\STSA}{STSA}
\icmltitlerunning{Structured Token Statistics Transformer}

\begin{document}
\twocolumn[
\icmltitle{Structured Token Statistics Transformer: Linear-Time Attention via Tighter Coding-Rate Bounds}
\begin{icmlauthorlist}
\icmlauthor{Menghui Zhou}{sheffield}
\icmlauthor{Po Yang}{sheffield}
\end{icmlauthorlist}
\icmlaffiliation{sheffield}{School of Computer Science, University of Sheffield, Sheffield, United Kingdom}
\icmlcorrespondingauthor{Menghui Zhou}{menghui.zhou@sheffield.ac.uk}
\icmlcorrespondingauthor{Po Yang}{po.yang@sheffield.ac.uk}
\icmlkeywords{white-box architectures, statistical attention, coding rate reduction}
\vskip 0.3in
]
\printAffiliationsAndNotice{}

\begin{abstract}
Token Statistics Transformer (ToST) obtains linear-time attention by summarizing projected tokens through coordinatewise second moments. This efficient statistic omits cross-coordinate moments: for fixed projections and memberships, different joint second-order structures can produce the same diagonal filter. Its coding-rate upper bound is tight only when the projected group moments are diagonal, a condition that a shared learned basis need not satisfy for each input. We introduce Structured Token Statistics Transformer (SToST), which retains joint moments within small feature-coordinate groups. Applying the Hadamard--Fischer inequality yields a coding-rate upper bound that is no looser than ToST's diagonal surrogate under matched projections and memberships, and strictly tighter when a retained within-group off-diagonal moment is nonzero. A gradient step produces Structured Token Statistics Attention (STSA), replacing scalar coordinate gates with small positive definite linear solves. Singleton groups recover the theoretical ToST operator. Group sizes may differ; when their maximum size is bounded independently of sequence length and head width, the statistical core retains ToST's $O(np)$ time and $O(p)$ statistics storage, with additional constant overhead. In single-run size-two-group pilots, SToST improves CIFAR-10 accuracy from 86.42\% to 86.81\% and reduces WikiText-2 test perplexity from 326.09 to 318.91, with higher measured training time in both settings. The resulting architecture preserves selected cross-coordinate statistics without introducing token-pair attention.
\end{abstract}

\section{Introduction}
\label{sec:introduction}

Self-attention allows tokens to exchange information across an input, but comparing every pair of tokens becomes costly as sequences grow \citep{vaswani2017attention}. This motivates a basic architectural question: how much information about an input must an attention layer retain to support useful global interaction? Token Statistics Transformer (ToST) offers one answer: summarize the tokens through compact statistics and use those statistics to adjust their features \citep{wu2025token}. This avoids a matrix of pairwise token interactions and gives an attention mechanism whose cost grows linearly with sequence length. ToST also follows the white-box approach developed through maximal coding rate reduction (MCR$^2$) and CRATE, in which network updates are derived from explicit representation objectives \citep{yu2020learning,yu2024whitebox}.

The efficiency of ToST comes with a specific statistical tradeoff. Its attention filter summarizes the energy of each projected feature separately, leaving out how different features vary together. For example, two features can have the same individual energies across two inputs, yet tend to activate together in one input and separately in the other. Their individual statistics do not reveal this difference. At the same feature projection and token assignments, ToST's diagonal filter therefore treats these joint structures alike. Learning the projection can make individual features more informative, but a shared projection need not remove these dependencies for every input. This motivates our question: \emph{can an attention layer retain selected relationships between features while preserving the efficiency of token statistics?}

We introduce \emph{Structured Token Statistics Transformer} (SToST), which summarizes small groups of features jointly. Within each attention head, we divide the feature coordinates into groups and retain both their individual energies and their pairwise cross moments. Each group still aggregates information across the tokens; grouping controls which feature relationships are preserved. The resulting \emph{Structured Token Statistics Attention} (STSA) operator adjusts the features within a group jointly. A group of two features already captures a relationship that separate scalar gates omit. Keeping the maximum group size fixed preserves the linear scaling of ToST's statistical core in both sequence length and head width, with additional computational overhead.

Our analysis connects this richer statistic to the objective underlying the attention update. ToST's diagonal approximation provides an upper bound on the projected coding-rate objective. Using a classical determinant inequality, we show that retaining within-group cross moments makes this bound no looser, and strictly tighter whenever any retained cross moment is nonzero. This comparison holds at the same projections and token assignments. We then derive STSA as a gradient step on the structured objective; groups containing a single feature recover ToST's theoretical attention operator. The method thus provides a controlled progression from separate feature statistics to joint statistics, together with an explicit account of the information retained and its computational cost.

This construction follows the principle of deriving learnable layers from optimization steps \citep{gregor2010learning,monga2021unrolling}. It also relates to architectures that model feature interactions, such as XCiT \citep{ali2021xcit}, and to DMSA's treatment of token memberships and feature subspaces within coding-rate models \citep{liu2026decoupled}. Our contribution is to identify and reduce the approximation made by ToST's diagonal statistical filter, and to derive an attention operator from that refinement. The determinant inequality is classical; its role here is to explain how retaining selected feature relationships changes the objective and the resulting layer.

Initial experiments on CIFAR-10 and WikiText-2 compare ToST with SToST using groups of two features. The single-run comparisons show modest improvements in classification accuracy and language-model perplexity, alongside higher measured training and inference costs. These results provide preliminary evidence for the proposed tradeoff, while a broader performance advantage remains to be established. A tighter coding-rate bound does not by itself guarantee better predictions: the practical question is whether the additional feature relationships justify their cost.

\section{From ToST to Structured Token Statistics}
\label{sec:method}

\subsection{Coding Rates and Projected Token Statistics}
\label{sec:setup}

Let $\Z=[\mathbf z_1,\ldots,\mathbf z_n]\in\R^{d\times n}$ contain tokens as columns. A row-stochastic matrix $\Pim\in\R_+^{n\times K}$ assigns each token to $K$ groups, with $\Pim\mathbf1=\mathbf1$. Write $\pik=\Pim_{:,k}$, $\mathbf P_k=\Diag(\pik)$, and $n_k=\mathbf1^\top\pik>0$. For $c=d/\epsilon^2>0$, the MCR$^2$ compression term is
\begin{align}
\M_k&=n_k^{-1}\Z\mathbf P_k\Z^\top,\label{eq:moment}\\
R_c(\Z,\Pim)&=\frac12\sum_{k=1}^K\frac{n_k}{n}
\log\det(\I_d+c\M_k).\label{eq:original}
\end{align}
These are uncentered second moments, not mean-subtracted covariances. The full rate-reduction objective is $R(\Z)-R_c(\Z,\Pim)$, where $R(\Z)=\frac12\log\det(\I_d+\frac cn\Z\Z^\top)$. Following ToST, we use the compression term to derive the attention sublayer; it is not proposed as a stand-alone training loss, whose minimization without further constraints could collapse the features.

For each head, let $\U_k\in\R^{d\times p}$ have orthonormal columns. Define
\begin{equation}
\A_k=\U_k^\top\Z,\qquad
\C_k=n_k^{-1}\A_k\mathbf P_k\A_k^\top.
\label{eq:projected-moment}
\end{equation}
The corresponding projected compression rate is
\begin{equation}
R_c^{\mathrm{proj}}=\frac12\sum_{k=1}^K\frac{n_k}{n}
\log\det(\I_p+c\C_k).
\label{eq:projected-rate}
\end{equation}
If $\range(\M_k)\subseteq\range(\U_k)$ for every $k$, then $R_c^{\mathrm{proj}}=R_c$: the projection removes only zero eigenvalues. Square orthogonal $\U_k$ satisfy coverage automatically. We first derive a bound on \eqref{eq:projected-rate}, then specify when it also bounds \eqref{eq:original}.

\subsection{ToST's Diagonal Approximation}
\label{sec:tost-limit}

At matched projections, memberships, and coding scale, ToST's theoretical compression surrogate is
\begin{equation}
G_{\mathrm{diag}}=\frac12\sum_k\frac{n_k}{n}
\sum_{i=1}^p\log(1+cC_{k,ii}).
\label{eq:tost-diag}
\end{equation}
Hadamard's inequality gives $R_c^{\mathrm{proj}}\leq G_{\mathrm{diag}}$, with equality precisely when every $\C_k$ is diagonal. Holding memberships fixed, its headwise filter is
\begin{equation}
\D_k^{\mathrm{diag}}=\Diag\!\left(
\frac{c}{1+cC_{k,11}},\ldots,\frac{c}{1+cC_{k,pp}}\right).
\label{eq:tost-filter}
\end{equation}
This exposes the computational tradeoff: $p$ scalar moments suffice, but the filter does not use $C_{k,ij}$ for $i\ne j$.

A two-dimensional example makes the limitation precise. Consider
\begin{equation}
\C_0=\I_2,\qquad
\C_\rho=\begin{bmatrix}1&\rho\\\rho&1\end{bmatrix},
\quad 0<|\rho|<1.
\label{eq:intro-example}
\end{equation}
Both yield the same diagonal statistic and the same TSSA gate. For a coding scale $c>0$, however, their full rates differ:
\begin{align}
r(\C_0)&=\log(1+c),\nonumber\\
r(\C_\rho)&=\tfrac12\log\big((1+c)^2-c^2\rho^2\big).
\label{eq:intro-rates}
\end{align}
The diagonal surrogate assigns $\log(1+c)$ to both and therefore overestimates the latter. The matrices with $+\rho$ and $-\rho$ also have the same diagonal but different principal directions. A coordinatewise gate does not distinguish these orientations, whereas a joint two-coordinate filter does. This is a limitation of the statistic under a common basis and assignment, not a claim that complete ToST networks produce identical outputs or cannot learn useful correlations.

ToST can learn a basis that exposes useful structure on the diagonal; however, its shared parameters do not enforce input-specific diagonalization. Our objective is to reduce the residual approximation in that same basis while keeping each statistical computation small.

\subsection{A Structured Bound That Tightens ToST}
\label{sec:bounds}

Let $\partn=\{B_1,\ldots,B_m\}$ partition the $p$ projected coordinates, with arbitrary positive sizes $b_r=|B_r|$ satisfying $\sum_r b_r=p$. These are groups of feature coordinates, not partitions of the token sequence. Set $\A_{k,r}=(\A_k)_{B_r,:}$ and $\U_{k,r}=(\U_k)_{:,B_r}$. The block statistic
\begin{equation}
\C_{k,r}=\frac1{n_k}\A_{k,r}\mathbf P_k\A_{k,r}^\top
=\sum_{j=1}^n\frac{\Pi_{jk}}{n_k}\mathbf a_{k,r,j}\mathbf a_{k,r,j}^\top
\label{eq:block-moment}
\end{equation}
is a $b_r\times b_r$ principal block of $\C_k$. It retains cross moments within $B_r$ without explicitly storing cross moments between blocks. Define
\begin{equation}
G_{\partn}(\Z,\Pim;\U)=\frac12\sum_{k=1}^K\frac{n_k}{n}
\sum_{r=1}^m\log\det(\I_{b_r}+c\C_{k,r}),
\label{eq:block-objective}
\end{equation}
where $\U$ denotes the collection of head projections.

\begin{proposition}[Blockwise hierarchy]
\label{prop:hierarchy}
For fixed $\Z$, $\Pim$, and $\U$, let $\partn_f$ refine $\partn_c$, meaning every fine block is contained in a coarse block. Then
\begin{equation}
R_c^{\mathrm{proj}}\leq G_{\partn_c}\leq G_{\partn_f}
\leq G_{\mathrm{diag}},
\label{eq:hierarchy}
\end{equation}
where $G_{\mathrm{diag}}$ uses singleton blocks. Equality $R_c^{\mathrm{proj}}=G_{\partn}$ holds if and only if each $\C_k$ is block diagonal with respect to $\partn$. Equality $G_{\partn}=G_{\mathrm{diag}}$ holds if and only if every retained block $\C_{k,r}$ is diagonal. Thus the improvement over ToST is strict whenever any retained within-block off-diagonal moment is nonzero.
\end{proposition}

\noindent The result follows from the Hadamard--Fischer inequality for positive definite matrices \citep{horn2013matrix}; a proof is given in \cref{app:hierarchy}. With coverage, $R_c$ replaces $R_c^{\mathrm{proj}}$ in \eqref{eq:hierarchy}. For square orthogonal projections, choosing an eigenbasis for every $\M_k$ attains equality, giving the variational identity
\begin{equation}
R_c(\Z,\Pim)=\min_{\U_1,\ldots,\U_K\in\mathrm O(d)}
G_{\partn}(\Z,\Pim;\U),\qquad p=d.
\label{eq:variational}
\end{equation}
No eigendecomposition of each input is required by the proposed layer: the projections are learned parameters. Consequently, a particular forward pass need not attain the minimum in \eqref{eq:variational}. In particular, structured statistics do not improve on a diagonal surrogate already optimized over a separate eigenbasis for every input. Their advantage is a tighter bound at the shared basis actually used by a layer. Bound comparisons require the same projections and memberships; they do not order the losses of independently trained models. The ordering between intermediate partitions requires refinement; two non-nested partitions need not be ordered.

\paragraph{How much of ToST's gap is removed?}
The improvement is exactly
\begin{equation}
G_{\mathrm{diag}}-G_{\partn}
=\frac12\sum_k\frac{n_k}{n}\sum_r
\log\frac{\prod_{i\in B_r}(1+cC_{k,ii})}
{\det(\I_{b_r}+c\C_{k,r})}.
\label{eq:improvement}
\end{equation}
Every summand is nonnegative. For a two-coordinate block $\C_{k,r}=\left[\begin{smallmatrix}a&\rho\\\rho&b\end{smallmatrix}\right]$, its contribution before the head weight is
\begin{equation}
-\frac12\log\left(1-\frac{c^2\rho^2}{(1+ca)(1+cb)}\right).
\label{eq:pair-gain}
\end{equation}
Thus a size-two block already yields a strict gain for $\rho\ne0$. It does not recover correlations outside that pair: block selection controls which part of the diagonal approximation is removed, and larger blocks incur greater cost.

\paragraph{What gap remains?}
Let $\mathbf H_k=\I_p+c\C_k$ and let $\mathbf T_k$ retain only its principal blocks in $\partn$. Then
\begin{align}
G_{\partn}-R_c^{\mathrm{proj}}
&=\sum_k\frac{n_k}{n}\,
\mathrm{KL}\!\left(\mathcal N(0,\mathbf H_k)
\,\Vert\,\mathcal N(0,\mathbf T_k)\right).
\label{eq:kl-gap}
\end{align}
Thus the gap quantifies discarded inter-block dependence in a noise-regularized Gaussian surrogate, not independence of the original tokens. Merging blocks retains more of this dependence. The identity and its assumptions are derived in \cref{app:gap}.

\subsection{From Scalar TSSA Gates to Structured Filters}
\label{sec:update}

We differentiate \eqref{eq:block-objective} with $\Pim$ and $\U$ fixed. For each head and block, define the positive definite resolvent
\begin{equation}
\begin{aligned}
\D_{k,r}&=c(\I_{b_r}+c\C_{k,r})^{-1},\\
\D_k&=\BlockDiag(\D_{k,1},\ldots,\D_{k,m}).
\end{aligned}
\label{eq:resolvent}
\end{equation}
The identity $d\log\det\mathbf H=\tr(\mathbf H^{-1}d\mathbf H)$ gives
\begin{equation}
\nabla_{\A_{k,r}}\!\left[
\frac{n_k}{2n}\log\det(\I_{b_r}+c\C_{k,r})\right]
=\frac1n\D_{k,r}\A_{k,r}\mathbf P_k.
\label{eq:block-gradient}
\end{equation}
The two moment factors cancel $1/2$, and the normalization cancels $n_k$. The chain rule yields
\begin{equation}
\nabla_\Z G_{\partn}
=\frac1n\sum_{k=1}^K\U_k\D_k\U_k^\top\Z\mathbf P_k.
\label{eq:gradient}
\end{equation}
We define STSA as the increment in a step of size $\tau>0$:
\begin{align}
\STSA_{\partn}(\Z;\Pim,\U)
&=-\frac\tau n\sum_{k=1}^K\U_k\D_k\U_k^\top\Z\mathbf P_k,
\label{eq:stsa}\\
\Z^+&=\Z+\STSA_{\partn}(\Z;\Pim,\U).
\label{eq:residual}
\end{align}
\Cref{app:gradient} provides the differential calculation. Global token interaction occurs through the memberships and aggregated block statistics; no $n\times n$ token-affinity matrix is formed.

\paragraph{Spectral interpretation.}
If $\C_{k,r}=\mathbf V\Lambda\mathbf V^\top$, then
\begin{equation}
\D_{k,r}=\mathbf V\Diag\!\left(
\frac{c}{1+c\lambda_1},\ldots,
\frac{c}{1+c\lambda_{b_r}}\right)\mathbf V^\top.
\label{eq:spectral}
\end{equation}
In a residual update, a high-energy direction therefore receives a smaller subtractive correction than a low-energy direction. Unlike a coordinatewise gate, this transformation can act along correlated directions inside a block. For example, the two matrices in \eqref{eq:intro-example} yield different resolvents despite producing the same diagonal gate. The resolvent is a spectral filter, not a matrix of attention probabilities; its off-diagonal entries need not be nonnegative.

\paragraph{Limits and invariances.}
For singleton blocks, \eqref{eq:resolvent} reduces to $c/(1+cC_{k,ii})$, recovering the theoretical TSSA gradient with the same memberships, projections, and step size. One block of size $p$ recovers the gradient of the full projected log-determinant. Intermediate partitions interpolate between these endpoints. In addition, replacing $\U_k$ by $\U_k\mathbf V_k$ for any block-diagonal orthogonal $\mathbf V_k$ leaves both $G_{\partn}$ and \eqref{eq:stsa} unchanged at fixed $\Pim$ (\cref{app:invariance}). The operator depends on each block's subspace, not on the particular orthonormal coordinates chosen within it.

\subsection{When Does the Bound Apply in the Original Space?}
\label{sec:coverage}

Small projections need not span every direction present in the input. In that case, \cref{prop:hierarchy} still bounds the projected rate but does not by itself upper-bound $R_c$. The missing energy can be accounted for explicitly.

\begin{proposition}[Residual-energy correction]
\label{prop:residual}
For arbitrary column-orthonormal $\U_k\in\R^{d\times p}$, define
\begin{equation}
E(\Z,\Pim;\U)=\frac{c}{2n}\sum_k
\big\|(\I_d-\U_k\U_k^\top)\Z\mathbf P_k^{1/2}\big\|_F^2.
\label{eq:residual-energy}
\end{equation}
Then $R_c\leq G_{\partn}+E\leq G_{\mathrm{diag}}+E$. Moreover, $E=0$ when the coverage condition holds, and, at fixed $\Pim,\U$,
\begin{equation}
\nabla_\Z E=\frac cn\sum_k(\I_d-\U_k\U_k^\top)\Z\mathbf P_k.
\label{eq:residual-gradient}
\end{equation}
\end{proposition}

\noindent The proof completes each projection to an orthogonal basis, applies Fischer's inequality, and bounds the omitted log-determinant by its trace (\cref{app:residual}). We distinguish two variants: \emph{projected STSA} uses \eqref{eq:stsa}, while \emph{coverage-corrected STSA} additionally subtracts $\tau\nabla_\Z E$. The corrected objective provides a general original-space upper bound without exact coverage, while preserving the improvement over the equivalently corrected diagonal objective. The correction can be evaluated without forming a dense $d\times d$ projector, using $\Z-\U_k(\U_k^\top\Z)$.

\subsection{Constructing SToST and Comparing Its Cost to ToST}
\label{sec:implementation}

\paragraph{Memberships and learning.}
We use the projected-energy routing rule from ToST's analytical construction:
\begin{equation}
\Pi_{jk}=\frac{\exp(\|\U_k^\top\mathbf z_j\|_2^2/(2\eta))}
{\sum_{h=1}^K\exp(\|\U_h^\top\mathbf z_j\|_2^2/(2\eta))},
\qquad\eta>0.
\label{eq:routing}
\end{equation}
Here $\eta$ is a routing temperature. With square orthogonal projections, all heads have the same projected energy and this rule gives uniform assignments. Distinct energy-based assignments therefore require proper subspaces or a different router. The input determines $\Pim$ first, after which \eqref{eq:gradient} is the \emph{partial} gradient holding those assignments fixed. It is not the total derivative of $G_{\partn}(\Z,\Pim(\Z);\U)$. End-to-end training may nevertheless backpropagate through the routing calculation and the linear solves: this outer differentiation trains the parameters of the explicit update. No detachment of the routing computation is implied by the partial-gradient derivation.

The reference layer uses fixed coordinate partitions, tied analysis/synthesis projections, and column-orthonormal $\U_k$. One implementation parameterizes $\U_k$ through the thin QR factor of a full-column-rank learnable matrix; the factorization is shared across inputs at fixed parameters and costs $O(dp^2)$ per head. Relaxing orthogonality or replacing the synthesis projection by an independent matrix defines a different architecture and does not inherit all the stated guarantees. ToST's practical implementation also normalizes projected feature rows across tokens for routing, absorbs a coding-scale coefficient into learned projections, and uses an independent output projection \citep[Appendix C.1]{wu2025token}. Our exact singleton recovery refers to its analytical TSSA operator with matched parameters. Reproducing the released layer requires matching those implementation choices as well. In particular, coordinatewise routing normalization need not preserve the within-block rotation invariance. Routing normalization, projection constraints, and parameter tying must be held constant when isolating the effect of group size.

A transformer-style use of the operator is
\begin{align}
\widetilde\Z^\ell&=\Z^\ell+
\STSA_{\partn}(\operatorname{LN}(\Z^\ell)),\label{eq:layer-a}\\
\Z^{\ell+1}&=\widetilde\Z^\ell+
\operatorname{MLP}(\operatorname{LN}(\widetilde\Z^\ell)),
\label{eq:layer-b}
\end{align}
where the STSA arguments include the layer's projections and the memberships computed from its normalized input. The gradient interpretation applies to the STSA increment on that input. Neither LayerNorm, the MLP, nor independently learned layers are asserted to implement descent on a single shared coding-rate objective.

\paragraph{Numerical realization.}
Rather than explicitly inverting \eqref{eq:resolvent}, solve
\begin{equation}
(\I_{b_r}+c\C_{k,r})\mathbf Y_{k,r}=c\A_{k,r}
\label{eq:solve}
\end{equation}
by Cholesky factorization. The identity term makes the system positive definite even for rank-deficient moments. If additional diagonal damping is introduced, it changes the operator and should be reported explicitly. \Cref{alg:stsa} specifies the projected variant.

\begin{algorithm}[t]
\caption{Projected STSA increment (one sample)}
\label{alg:stsa}
\begin{algorithmic}[1]
\REQUIRE $\Z$, projections $\{\U_k\}$, partition $\partn$, $c,\eta,\tau>0$
\STATE Compute $\A_k=\U_k^\top\Z$ for all heads.
\STATE Compute $\Pim$ by \eqref{eq:routing}; set $n_k=\sum_j\Pi_{jk}$.
\STATE Initialize $\Delta\Z=0$.
\FOR{$k=1,\ldots,K$}
  \FOR{$r=1,\ldots,m$}
    \STATE Form $\C_{k,r}=\A_{k,r}\mathbf P_k\A_{k,r}^\top/n_k$.
    \STATE Solve \eqref{eq:solve} for $\mathbf Y_{k,r}$.
    \STATE $\Delta\Z\leftarrow\Delta\Z-\frac\tau n\U_{k,r}\mathbf Y_{k,r}\mathbf P_k$.
  \ENDFOR
\ENDFOR
\STATE \textbf{return} $\Delta\Z$
\end{algorithmic}
\end{algorithm}

\paragraph{Complexity for unequal group sizes.}
For one head, forming and applying the $r$th block costs $O(nb_r^2)$, its Cholesky factorization costs $O(b_r^3)$, and its matrix storage is $O(b_r^2)$. Hence the statistical core requires
\begin{equation}
\begin{aligned}
T_{\mathrm{stat}}&=O\!\left(n\sum_r b_r^2+\sum_r b_r^3\right),\\
M_{\mathrm{stat}}&=O\!\left(\sum_r b_r^2\right).
\end{aligned}
\label{eq:general-complexity}
\end{equation}
The block sizes need not be equal. If $b_r\leq B$, then $\sum_r b_r^2\leq Bp$ and $\sum_r b_r^3\leq B^2p$, so
\begin{equation}
T_{\mathrm{stat}}=O(npB+pB^2),\qquad
M_{\mathrm{stat}}=O(pB).
\label{eq:bounded-complexity}
\end{equation}
For fixed $B$ and $n\geq1$, these reduce to TSSA's $O(np)$ and $O(p)$ orders. This statement concerns both sequence length and head width in the statistical core. Linear scaling in $n$ alone is a weaker property: even a full $p\times p$ moment has linear dependence on $n$ when $p$ is fixed.

Including analysis/synthesis projections, the $K$-head forward cost is
\begin{equation}
O\!\left(Kndp+KnpB+KpB^2\right),
\label{eq:complexity}
\end{equation}
plus the projection-parameterization cost. A direct implementation stores $O(nd+Knp+nK)$ token activations and memberships, in addition to $O(KpB)$ statistical matrices and $O(Kdp)$ projection parameters. The full layer therefore uses memory linear in $n$ at fixed model dimensions; only its statistical storage is independent of $n$. Standard reverse-mode differentiation through these operations has the same asymptotic orders. LayerNorm and a tokenwise MLP also preserve linear dependence on $n$ at fixed widths.

\paragraph{What does group size two add?}
For even $p$ and $b_r=2$, there are $p/2$ blocks. The sums in \eqref{eq:general-complexity} become $2p$ and $4p$, respectively, giving $O(2np+4p)$ core arithmetic and $2p$ dense matrix entries, compared with $p$ scalar entries for TSSA. Exploiting symmetry reduces the stored moment entries to $3p/2$. These counts describe extra statistical work, not an exact factor-of-two runtime prediction: diagonal gates and small solves have different constants and hardware utilization. For odd $p$, one singleton completes the partition.

\begin{table}[t]
\caption{Per-head statistical core. Uniform-size rows assume $b\mid p$. Projections and token activations are excluded.}
\label{tab:complexity}
\centering
\begin{tabular}{lcc}
\toprule
Statistic & Arithmetic & Matrix storage\\
\midrule
ToST ($b=1$) & $O(np)$ & $O(p)$\\
STSA (size $b$) & $O(npb+pb^2)$ & $O(pb)$\\
STSA ($b_r\leq B$) & $O(npB+pB^2)$ & $O(pB)$\\
Full ($b=p$) & $O(np^2+p^3)$ & $O(p^2)$\\
\bottomrule
\end{tabular}
\end{table}

Fixing the \emph{number} of equal blocks $m$ instead yields $b=p/m$ and $O(np^2/m+p^3/m^2)$ core arithmetic. It retains linear scaling in $n$ but not TSSA's linear scaling in $p$. The design choice behind our efficiency claim is bounded block size, which also permits unequal sizes without changing the order.

\paragraph{Scope of the optimization claim.}
For fixed assignments and projections, a sufficiently small gradient step decreases the smooth surrogate. A decrease of an upper bound alone need not decrease $R_c$: a majorization argument additionally requires the bound to touch $R_c$ at the current iterate. A tighter bound likewise does not imply a higher task accuracy. STSA makes the retained statistical information and its computational price explicit; its downstream benefits remain an empirical question.

\section{Preliminary Experiments}
\label{sec:experiments}

We compare ToST's coordinatewise statistical filter with SToST's size-two feature groups in image classification and causal language modeling. The intervention replaces the attention operator; model dimensions, initial weights, and training recipes are matched within each task. A size-two group retains one cross-coordinate second moment and applies a $2\times2$ filter. Each configuration was trained once with seed 42. These are initial task-level comparisons, not reproductions of ToST's published ImageNet-1K or OpenWebText results.

\paragraph{CIFAR-10 classification.}
We adapt the official 12-layer ToST-Tiny backbone to $32\times32$ images using the same two-convolution, stride-four patch stem for both models, producing 64 tokens. Both use width 192, four heads, an MLP expansion ratio of four, and a ten-class classifier, with 5.53M parameters. We train on the 50,000-image training split for 50 epochs using AdamW, batch size 128, a five-epoch warmup, cosine learning-rate decay from $5\times10^{-4}$, and standard random crop and horizontal flip augmentation. We evaluate the final checkpoint on all 10,000 test images; the test set is not used for checkpoint selection. Both runs start from identical parameters and use the same seed and hyperparameters.

\begin{table}[t]
\centering
\caption{CIFAR-10 pilot: test accuracy, training time, and full-model inference latency.}
\label{tab:cifar-pilot}
\footnotesize
\setlength{\tabcolsep}{4pt}
\begin{tabular}{lrrr}
\toprule
Model & Acc. (\%) & Train (min) & Infer. (ms) \\
\midrule
ToST-Tiny & 86.42 & \textbf{17.14} & \textbf{15.36} \\
SToST-Tiny, $B=2$ & \textbf{86.81} & 34.72 & 22.14 \\
\bottomrule
\end{tabular}
\end{table}

As shown in \cref{tab:cifar-pilot}, SToST improves test accuracy by 0.39 percentage points, while its measured training time is $2.03\times$ that of ToST. Training time sums epoch wall times. Inference is a full-model forward pass on resident GPU tensors with batch size 128, after 10 warmup and 100 timed batches. The two runs used different machines with the same RTX 4060 Ti GPU model; this pilot does not isolate attention-kernel overhead from host-level timing differences.

\paragraph{WikiText-2 causal language modeling.}
We keep the official ToST-Base language architecture: 12 layers, width 768, 12 heads, context length 1024, and a tied GPT-2 embedding/output vocabulary of 50,304 entries (110.38M parameters). Both models are initialized identically. We tokenize the raw WikiText-2 splits with the GPT-2 tokenizer, train from scratch on the 2.40M-token training split for 20 epochs, and use AdamW with global batch size 32, peak learning rate $6\times10^{-4}$, and cosine decay. Each model takes 1,480 optimizer steps and processes 48.07M training tokens. We select the checkpoint with the lowest validation loss and evaluate it once on the complete 284,614-target test split. Perplexity is computed over GPT-2 subword tokens. Each run used one A100-SXM4-80GB GPU on a different HPC node.

\begin{table}[t]
\centering
\caption{WikiText-2 pilot: test perplexity, training time, and teacher-forced evaluation throughput.}
\label{tab:wikitext-pilot}
\footnotesize
\setlength{\tabcolsep}{4pt}
\begin{tabular}{lrrr}
\toprule
Model & PPL & Train (min) & Eval. tok/s \\
\midrule
ToST-Base & 326.09 & \textbf{15.26} & \textbf{145,559} \\
SToST-Base, $B=2$ & \textbf{318.91} & 22.59 & 96,645 \\
\bottomrule
\end{tabular}
\end{table}

Both models select epoch 7 on the validation split. SToST lowers test perplexity by 7.18 (2.20\%) at $1.48\times$ the pure training time of ToST; its teacher-forced test evaluation is $1.51\times$ slower. Training time accumulates synchronized optimizer-update durations, excluding validation and checkpoint writing. Evaluation throughput includes data transfer and full output logits; it does not measure cached autoregressive generation. Training from scratch on WikiText-2 is a different protocol from ToST's OpenWebText pretraining and zero-shot evaluation, so these perplexities should not be compared with the published language results. Together, these single-run pilots suggest a favorable quality change under size-two grouping on the two tasks, alongside a substantial measured runtime cost. They do not establish statistical reliability or a general accuracy advantage.

\section{Discussion and Conclusion}
\label{sec:discussion}

ToST demonstrates that global token interaction can be built from compact statistics. Our analysis identifies a specific cost of its diagonal approximation: at a fixed learned basis, cross-coordinate second moments do not enter the statistical gate, even when they account for a nonzero coding-rate gap. SToST retains selected joint moments and yields a bound that is strictly tighter whenever those retained correlations are nonzero. The resulting STSA operator replaces scalar gates with small matrix filters. Bounded group sizes preserve the asymptotic cost of ToST's statistical core while allowing unequal allocations of statistical capacity.

This refinement has limits. Correlations between distinct groups are still omitted, and the result does not identify which partition is best for a task. A basis that already diagonalizes the moments leaves no gap to remove. Fixed projections and memberships are essential to the pointwise comparison; independently trained models need not retain that ordering. The original-space interpretation additionally requires subspace coverage or the residual correction. These conditions specify what the tighter-bound result establishes without attributing a universal representational failure to ToST.

The CIFAR-10 and WikiText-2 pilots show modest quality gains for size-two groups under the tested recipes, with measurable training and inference overhead. These single runs do not establish a general performance advantage, and the CIFAR-10 timings include machine-level variation. Full ImageNet-1K and OpenWebText comparisons remain to be conducted. Further evaluation should isolate the retained statistic across singleton, size-two, and larger bounded groups under matched routing and projection choices, and report the coding-rate gap alongside task quality, latency, and peak memory. A tighter surrogate and the same asymptotic complexity do not guarantee improved accuracy or equal runtime. The central design principle is to preserve a controlled amount of joint feature structure within token-statistics attention.

\bibliography{references}
\bibliographystyle{icml2027}

\clearpage
\appendix
\onecolumn
\section{Supporting Derivations}
\label{app:proofs}

\subsection{Tightening the ToST Bound: Hierarchy and Equality}
\label{app:hierarchy}

For a positive definite matrix $\mathbf H$ with two principal blocks, the Schur-complement identity gives
\begin{equation}
\det\mathbf H=\det\mathbf H_{11}
\det\!\left(\mathbf H_{22}-\mathbf H_{21}\mathbf H_{11}^{-1}\mathbf H_{12}\right)
\leq\det\mathbf H_{11}\det\mathbf H_{22}.
\end{equation}
The inequality follows because $\mathbf H_{21}\mathbf H_{11}^{-1}\mathbf H_{12}$ is positive semidefinite and $\log\det$ is increasing in the positive definite order. Equality holds precisely when $\mathbf H_{12}=0$. Applying this argument recursively to a partition yields the Hadamard--Fischer inequality and its equality condition: $\det\mathbf H\leq\prod_r\det\mathbf H_{B_rB_r}$, with equality if and only if $\mathbf H$ is block diagonal for that partition.

Take $\mathbf H_k=\I_p+c\C_k\succ0$. Taking logarithms and then summing with positive weights $n_k/(2n)$ proves $R_c^{\mathrm{proj}}\leq G_{\partn}$. Applying the same argument inside each coarse block proves $G_{\partn_c}\leq G_{\partn_f}$ for nested partitions; refining to singletons gives the final bound in \eqref{eq:hierarchy}. Positivity of the weights means equality of the aggregate requires equality for every head. Because $c>0$, the off-block entries of $\mathbf H_k$ vanish exactly when those of $\C_k$ do. Applying the singleton equality condition separately to each retained block proves $G_{\partn}=G_{\mathrm{diag}}$ if and only if every $\C_{k,r}$ is diagonal. A single nonzero within-block off-diagonal entry makes that block's inequality strict and, because $n_k>0$, makes the aggregate improvement strict. None of these arguments requires equal block sizes.

For a column-orthonormal $\U_k$ covering $\range(\M_k)$, complete it to an orthogonal matrix $[\U_k,\mathbf V_k]$. Then
\begin{equation}
[\U_k,\mathbf V_k]^\top\M_k[\U_k,\mathbf V_k]
=\begin{bmatrix}\C_k&0\\0&0\end{bmatrix},
\end{equation}
so $\det(\I_d+c\M_k)=\det(\I_p+c\C_k)$. When $p=d$, an orthogonal eigenbasis of each $\M_k$ makes $\C_k$ diagonal and hence block diagonal for every partition. This attains the lower bound and proves \eqref{eq:variational}. The same argument applies to a rank-restricted basis when the coverage condition is feasible.

\subsection{The Removed and Remaining Diagonal Approximation Gap}
\label{app:gap}

Subtracting \eqref{eq:block-objective} from \eqref{eq:tost-diag} and grouping the scalar terms by $B_r$ gives \eqref{eq:improvement}. Hadamard's inequality makes every log ratio nonnegative. In the two-dimensional case,
\begin{equation}
\det(\I_2+c\C_{k,r})=(1+ca)(1+cb)-c^2\rho^2,
\end{equation}
which gives \eqref{eq:pair-gain} after division by $(1+ca)(1+cb)$. Positive semidefiniteness implies $a,b\geq0$ and $\rho^2\leq ab$, so the ratio inside the logarithm is strictly positive. The gain is zero exactly when $\rho=0$. This removed gap complements the remaining gap below.

Let $\mathbf T_k=\BlockDiag((\mathbf H_k)_{B_1B_1},\ldots,(\mathbf H_k)_{B_mB_m})$. Both matrices are positive definite. The Gaussian divergence formula gives
\begin{equation}
\mathrm{KL}\!\left(\mathcal N(0,\mathbf H_k)\Vert\mathcal N(0,\mathbf T_k)\right)
=\frac12\left[\tr(\mathbf T_k^{-1}\mathbf H_k)-p+
\log\det\mathbf T_k-\log\det\mathbf H_k\right].
\end{equation}
Because $\mathbf T_k$ consists of exactly the principal blocks of $\mathbf H_k$, each diagonal block of $\mathbf T_k^{-1}\mathbf H_k$ is an identity matrix, and its trace is $p$. Also $\log\det\mathbf T_k=\sum_r\log\det(\I_{b_r}+c\C_{k,r})$. Multiplication by $n_k/n$ and summation prove \eqref{eq:kl-gap}. The reference Gaussian with covariance $\mathbf T_k$ is the product of the corresponding block marginals of the Gaussian with covariance $\mathbf H_k$. This probabilistic interpretation concerns the auxiliary Gaussian only; no Gaussian assumption about the actual tokens is needed for the determinant bound.

\subsection{Derivative of the Structured Objective}
\label{app:gradient}

Fix $k,r$ and suppress their indices. Set $\A=\U_{k,r}^\top\Z$, $\mathbf P=\mathbf P_k$, $\C=\A\mathbf P\A^\top/n_k$, and $\mathbf H=\I+c\C$. With memberships and projections fixed,
\begin{align}
dg
&=d\left[\frac{n_k}{2n}\log\det\mathbf H\right]
=\frac{c}{2n}\tr\!\left[\mathbf H^{-1}
(d\A\,\mathbf P\A^\top+\A\mathbf P\,d\A^\top)\right]\\
&=\frac1n\tr\!\left[(c\mathbf H^{-1}\A\mathbf P)^\top d\A\right].
\end{align}
Symmetry of $\mathbf P$ and $\mathbf H^{-1}$ makes the two trace terms equal. Hence $\nabla_\A g=c\mathbf H^{-1}\A\mathbf P/n=\D\A\mathbf P/n$. Substituting $d\A=\U_{k,r}^\top d\Z$ and summing over blocks and heads proves \eqref{eq:gradient}. This derivative calculation remains algebraically valid for an arbitrary fixed projection matrix; orthogonality is needed for the original-space variational interpretation, not for differentiation of the projected objective.

\subsection{Within-Block Orthogonal Invariance}
\label{app:invariance}

Let $\mathbf V_k=\BlockDiag(\mathbf V_{k,1},\ldots,\mathbf V_{k,m})$, where each $\mathbf V_{k,r}$ is orthogonal, and replace $\U_k$ by $\U_k'=\U_k\mathbf V_k$. At fixed memberships,
\begin{equation}
\A_{k,r}'=\mathbf V_{k,r}^\top\A_{k,r},\qquad
\C_{k,r}'=\mathbf V_{k,r}^\top\C_{k,r}\mathbf V_{k,r},\qquad
\D_{k,r}'=\mathbf V_{k,r}^\top\D_{k,r}\mathbf V_{k,r}.
\end{equation}
Orthogonal similarity preserves every block log-determinant. Moreover,
\begin{equation}
\U_{k,r}'\D_{k,r}'(\U_{k,r}')^\top
=\U_{k,r}\D_{k,r}\U_{k,r}^\top.
\end{equation}
Thus both the objective and the update are unchanged. The routing rule in \eqref{eq:routing} also preserves memberships under these transformations because the full projected squared norm is invariant. For a single full block, the invariance extends to all orthogonal coordinate changes inside the projected subspace. Arbitrary coordinatewise routing normalizations need not preserve this property and are not part of the reference construction.

\subsection{Residual-Energy Correction}
\label{app:residual}

Complete $\U_k$ to an orthogonal basis $\mathbf Q_k=[\U_k,\mathbf V_k]$ without assuming coverage. Write
\begin{equation}
\mathbf Q_k^\top\M_k\mathbf Q_k
=\begin{bmatrix}\C_k&\mathbf B_k\\\mathbf B_k^\top&\mathbf E_k\end{bmatrix},
\qquad\mathbf E_k=\mathbf V_k^\top\M_k\mathbf V_k\succeq0.
\end{equation}
Fischer's inequality and the scalar inequality $\log(1+x)\leq x$ for $x\geq0$ imply
\begin{align}
\log\det(\I_d+c\M_k)
&\leq\log\det(\I_p+c\C_k)+\log\det(\I_{d-p}+c\mathbf E_k)\\
&\leq\sum_r\log\det(\I_{b_r}+c\C_{k,r})+c\tr(\mathbf E_k).
\end{align}
Since $\mathbf V_k\mathbf V_k^\top=\I_d-\U_k\U_k^\top$, the residual trace is
\begin{equation}
\tr(\mathbf E_k)=\frac1{n_k}
\big\|(\I_d-\U_k\U_k^\top)\Z\mathbf P_k^{1/2}\big\|_F^2.
\end{equation}
Multiplication by $n_k/(2n)$ and summation give \cref{prop:residual}. The residual vanishes under coverage. Differentiating the quadratic form, using symmetry and idempotence of $\I_d-\U_k\U_k^\top$, yields \eqref{eq:residual-gradient}. If $p=d$, the residual is identically zero. Adding the same $E$ to $G_{\partn}\leq G_{\mathrm{diag}}$ proves the final comparison in \cref{prop:residual}; the correction does not alter the difference between the structured and diagonal objectives.

\subsection{Complexity with Bounded, Unequal Block Sizes}
\label{app:complexity}

Weighted moment aggregation multiplies a $b_r\times n$ matrix by an $n\times b_r$ matrix, costing $O(nb_r^2)$; multiplying token columns by membership weights costs $O(nb_r)$. Factoring a positive definite $b_r\times b_r$ system costs $O(b_r^3)$, and its triangular solves for $n$ right-hand sides cost $O(nb_r^2)$. Summing over blocks proves \eqref{eq:general-complexity}. The matrices and their factors require $O(\sum_r b_r^2)$ storage when retained for differentiation. With $b_r\leq B$ and $\sum_r b_r=p$,
\begin{equation}
\sum_r b_r^2\leq\sum_r Bb_r=Bp,
\qquad
\sum_r b_r^3\leq\sum_r B^2b_r=B^2p.
\end{equation}
These inequalities prove \eqref{eq:bounded-complexity} without a divisibility or equal-size assumption. Summed analysis/synthesis projection work is $O(nd\sum_r b_r)=O(ndp)$ per head. Projected-energy routing costs $O(Knp)$ for squared norms and $O(nK)$ for the softmax; both are absorbed in \eqref{eq:complexity}. Gradients of the products and solves use operations of the same orders. This counts dense arithmetic and storage, not wall-clock latency or a particular kernel implementation.

\subsection{Descent and Its Limits}
\label{app:descent}

For fixed $\Pim$ and $\U$, let $G$ denote either $G_{\partn}$ or $G_{\partn}+E$. These objectives are smooth because all log-determinant arguments include a positive identity term. If the gradient is $L$-Lipschitz on a region containing a step, the descent lemma gives
\begin{equation}
G(\Z-\tau\nabla G(\Z))
\leq G(\Z)-\tau\left(1-\frac{L\tau}{2}\right)\|\nabla G(\Z)\|_F^2.
\end{equation}
Thus $0<\tau<2/L$ gives strict descent away from a stationary point under this condition. This is a fixed-surrogate statement. For a majorization argument on $R_c$, one additionally needs an upper bound $G_t$ satisfying $G_t(\Z_t)=R_c(\Z_t,\Pim)$ and $G_t(\Z_{t+1})\leq G_t(\Z_t)$. Only then does
\begin{equation}
R_c(\Z_{t+1},\Pim)\leq G_t(\Z_{t+1})\leq G_t(\Z_t)=R_c(\Z_t,\Pim)
\end{equation}
follow. Re-estimating memberships, changing projection parameters across layers, or applying the surrounding normalization and MLP requires separate analysis. These facts delimit the optimization interpretation of the proposed attention increment.

\end{document}